\documentclass{article} %
\usepackage[T1]{fontenc}
\usepackage{iclr2027_conference,times}

\usepackage{amsmath,amsfonts,bm}

\def\eqref#1{equation~\ref{#1}}

\def\1{\bm{1}}

\DeclareMathAlphabet{\mathsfit}{\encodingdefault}{\sfdefault}{m}{sl}
\SetMathAlphabet{\mathsfit}{bold}{\encodingdefault}{\sfdefault}{bx}{n}

\usepackage{hyperref}
\usepackage{url}
\usepackage{graphicx}
\usepackage{xcolor}
\usepackage{colortbl}
\definecolor{appletransfer}{HTML}{E5EEF8}
\usepackage{booktabs}
\usepackage{tabularx}
\usepackage{wrapfig}
\usepackage{needspace}
\usepackage{amsmath}
\usepackage{amssymb}

\title{Bridging Frontier Reasoning and Robot Execution: \\
From Autonomous Demonstration Generation to Dense Language Supervision}

\AddToHook{begindocument/before}{%
  \iclrfinalcopy
  \author{%
    Bosung Kim\textsuperscript{1}\quad
    Alexander Trevithick\textsuperscript{2}\quad
    Ruiyi Wang\textsuperscript{1}\quad
    Prithviraj Ammanabrolu\textsuperscript{1}\\
    \makebox[\dimexpr\textwidth-2\tabcolsep\relax][c]{%
      \normalfont\textsuperscript{1}University of California, San Diego\qquad
      \textsuperscript{2}NVIDIA
    }
  }
  \hypersetup{%
    pdftitle={Bridging Frontier Reasoning and Robot Execution: From Autonomous Demonstration Generation to Dense Language Supervision},
    pdfauthor={Bosung Kim, Alexander Trevithick, Ruiyi Wang, Prithviraj Ammanabrolu}
  }
  \let\arxivmaketitle\maketitle
  \renewcommand{\maketitle}{%
    \arxivmaketitle
    \lhead{}
    \renewcommand{\headrulewidth}{0pt}%
  }
}

\begin{document}

\maketitle

\begin{abstract}
Recent advances in frontier models enable robot manipulation from only a few demonstrations, but high inference latency limits their use for real-time robot control. To bridge this gap, we study two complementary approaches that connect frontier reasoning with low-latency local execution. First, we use a frontier model to autonomously generate demonstrations that supplement human demonstrations for training a fast local policy. We augment its in-context examples with corrective demonstration segments that show how to recover from physical errors, improving generation reliability. Generation time and cost decrease as successful examples accumulate in context, suggesting a path toward more efficient data collection. At deployment, a harness combines frontier-generated instructions with a fast local policy, enabling efficient execution while preserving the frontier model's ability to guide and correct actions.

With low-latency execution delegated to the local policy, the bottleneck shifts to its capacity to reliably follow the frontier model's diverse instructions. Our second bridge introduces dense language supervision across three nested granularities---primitive, atomic, and composite---with multi-aspect descriptions at each level. Across long-horizon tasks in RoboCasa 365 and BEHAVIOR-1K, where instructions change as execution progresses, the combined supervision achieves the highest performance under both oracle and frontier-model instructors, demonstrating more reliable instruction following through the policy's language interface. Finally, we evaluate both bridges together on a crossword task that combines semantic planning and manipulation within a fixed time budget. These results support autonomous demonstration generation and dense language supervision as complementary components for connecting frontier reasoning to low-latency local execution.

\end{abstract}

\section{Introduction}
Robot learning has explored several approaches to translating frontier-level intelligence into physical robot actions. These approaches differ in how frontier models guide execution: they can issue instructions to a trained local policy~\citep{shi2025hirobot}, generate executable programs over control primitives~\citep{liang2022codeaspolicies}, provide spatial guidance such as trajectories~\citep{li2025hamster}, or produce robot commands directly~\citep{robocurve2026astra}. Recent advances in frontier models enable robot manipulation from only a few demonstrations through in-context learning, without task-specific fine-tuning. However, frontier-model inference and API latency limit the responsiveness of direct robot control, whereas practical manipulation requires high-frequency sensory feedback and timely action adjustments for reliable physical interaction~\citep{berman2026claudeplays}. This motivates our central question: how can we retain the benefits of frontier-model reasoning while delegating motor execution to a fast local policy?

The ability to manipulate objects from a few examples makes frontier models a potential source of additional robot demonstrations, complementing teleoperation-based data collection. Our first bridge explores this capability by generating manipulation trajectories and incorporating them into local-policy training. Successful demonstrations alone do not reliably support autonomous generation, particularly during precise contact operations. We augment the in-context examples with corrective demonstration segments: short demonstrations of adjustments at moments when execution encounters physical errors. As successful trajectories accumulate, we retain useful segments for reuse in subsequent episodes and examine how generation time and API cost evolve. At deployment, a harness combines frontier-generated plans with the policy's low-latency execution, allowing the frontier model to provide high-level direction while the policy generates motor actions within each execution segment.

Delegating motor execution reduces the need for frontier inference to generate actions, but makes the policy's instruction-following capabilities a central constraint. We propose dense language supervision as the next bridge, enabling the policy to interpret and execute diverse, fine-grained instructions generated by the frontier model. We annotate the same demonstrations at three nested granularities---primitive, atomic, and composite---and pair these annotations with multi-aspect descriptions to broaden instruction diversity.

We study two complementary bridges between frontier reasoning and robot execution: generating demonstrations for local-policy training and expanding the policy's language interface through dense supervision. Across four manipulation tasks, mixed human and generated demonstrations achieve the same aggregate policy success as an equally sized human-only corpus, supporting generated trajectories as a supplementary training source. Hierarchical and multi-aspect supervision together achieve the strongest results across both simulation benchmarks and instructor conditions. Finally, the crossword task demonstrates their integration, with dense supervision improving performance under dynamic spatial instructions. Together, these results support autonomous demonstration generation and dense language supervision as complementary bridges between frontier reasoning and low-latency local execution.

\section{Related Work}

\textbf{Foundation Models for Robot Control.} Foundation models guide robot execution through several interfaces, summarized in Figure~\ref{fig:route-taxonomy} in Appendix~\ref{app:control-taxonomy}. Direct-control evaluations of frontier models demonstrate manipulation capabilities alongside limitations in execution speed and precision~\citep{robocurve2026astra,berman2026claudeplays}. Hierarchical and dual-system architectures couple vision-language processing with low-level control~\citep{bjorck2025groot,shi2025hirobot,chen2026volo,cheng2024navila}, while other approaches provide spatial guidance~\citep{li2025hamster,shridhar2024genima,lee2026mu0} or synthesized goal images~\citep{black2023susie}. Hi Robot fine-tunes a high-level VLM using labeled demonstrations and synthetic language interactions, and a low-level VLA using robot demonstrations~\citep{shi2025hirobot}. Harness frameworks such as Thea support feedback-driven task execution~\citep{wang2026thea}. Our study retains a frontier model without task-specific fine-tuning and examines how to train a local policy that executes its instructions.

\textbf{Autonomous Data Generation and Policy Learning.} MimicGen adapts human demonstrations to new scene configurations and embodiments~\citep{mandlekar2023mimicgen}. Manipulate-Anything uses VLM-generated actions and visual verification to collect physical demonstrations for policy learning~\citep{duan2024manipulate}. Show-Harness collects human and frontier-generated demonstrations through predefined semantic actions mapped to robot commands by an embodiment-specific interpreter, and uses the resulting trajectories for policy training~\citep{chen2026showharness}. Program-synthesis approaches generate executable programs over perception and control primitives~\citep{liang2022codeaspolicies,fu2026capx}, with ASPIRE accumulating reusable skills~\citep{lu2026aspire} and ENPIRE extending agent-driven improvement to neural policies, training infrastructure, and algorithms~\citep{xiao2026enpire}. Our first bridge studies autonomous demonstration generation through direct, low-level robot control by frontier models. Using successful demonstrations and corrective demonstration segments as in-context references, a frontier model executes tasks with visual feedback and produces new training trajectories. We examine generation reliability, track how generation time and cost evolve as successful examples accumulate, and evaluate the resulting demonstrations as a supplementary source for local-policy training.

\textbf{Language Supervision and Instruction Following.} RT-H learns language-motion prediction and action generation within a single VLM, enabling corrections through language~\citep{belkhale2024rth}. \citet{hu2026orchestrating} show that a controller's responsiveness to varied instructions affects hierarchical execution. DeMiAn provides multi-aspect segment annotations and a learned instructor, including evaluation on composite tasks with changing atomic instructions~\citep{demian2026}. Building on multi-aspect supervision, our second bridge jointly annotates shared trajectories at three nested granularities---primitive, atomic, and composite---and adds alternative descriptions and rephrasings. We examine how temporal granularity and linguistic diversity affect instruction following under oracle and frontier-model instructors, with additional robustness evaluations across planners and instruction phrasings.

\section{Bridge 1: Distilling Frontier Reasoning into Local Control}
\label{sec:bridge1}

Our first bridge uses frontier-generated demonstrations to supplement human demonstrations for training a fast local policy. We first examine autonomous generation reliability, then generation efficiency, and finally deployment through a planner--policy harness.

\subsection{Autonomous Rollouts via In-Context Learning}
\label{sec:autonomous-rollouts}

We first evaluate the capacity of frontier models to autonomously control a robot given only a minimal set of human demonstrations.
We benchmark two models, Fable 5 and GPT-6 Astra, across four tasks spanning a spectrum of physical complexity: basic pick-and-place (claw machine), spatial alignment (stacking cups), deformable manipulation (folding a cloth), and high-precision contact (closing a lid).
Each trial is provided five human demonstrations with zero subsequent human intervention, and the environment is fully reset between attempts.
Trials exceeding a one-hour timeout are recorded as failures. This limit applies to the capability evaluation; the subsequent data-collection runs use a separate protocol.

\begin{figure}[t]
\centering
\includegraphics[width=\linewidth,height=0.72\textheight,keepaspectratio]{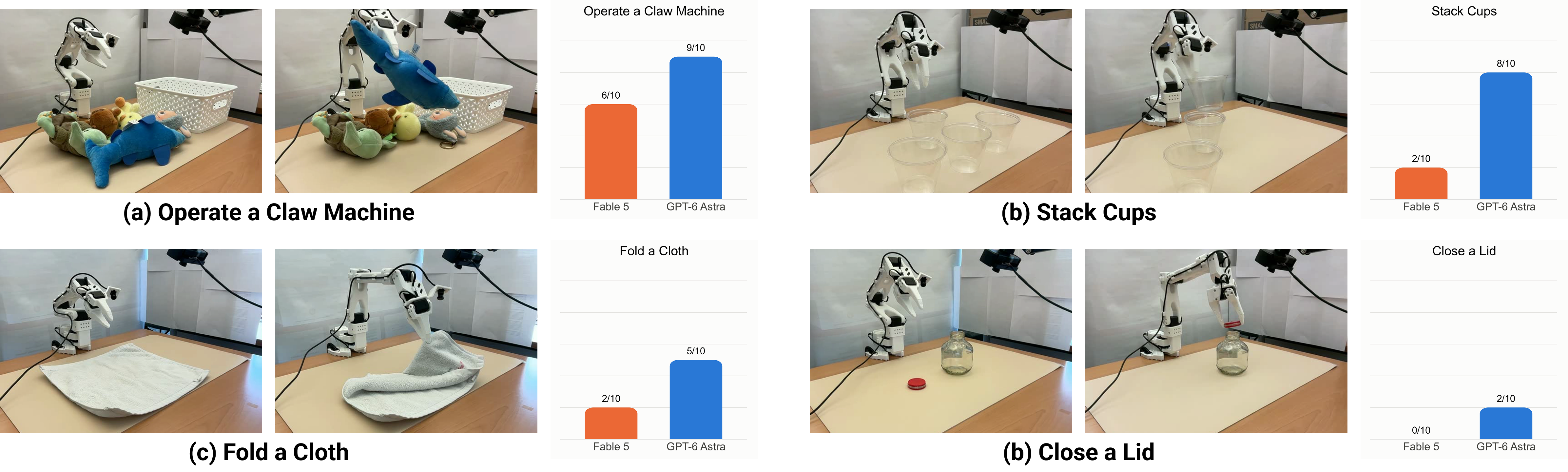}
\caption{Autonomous manipulation from five successful in-context demonstrations. Task views and success counts are shown for Fable 5 and GPT-6 Astra across four tasks, with ten trials per model and task. Each trial starts from a reset environment and has a one-hour timeout, with no subsequent human intervention.}
\label{fig:bridge1-2}
\end{figure}

\Needspace{18\baselineskip}
\begin{wrapfigure}{r}{0.40\textwidth}
\centering
\includegraphics[width=\linewidth]{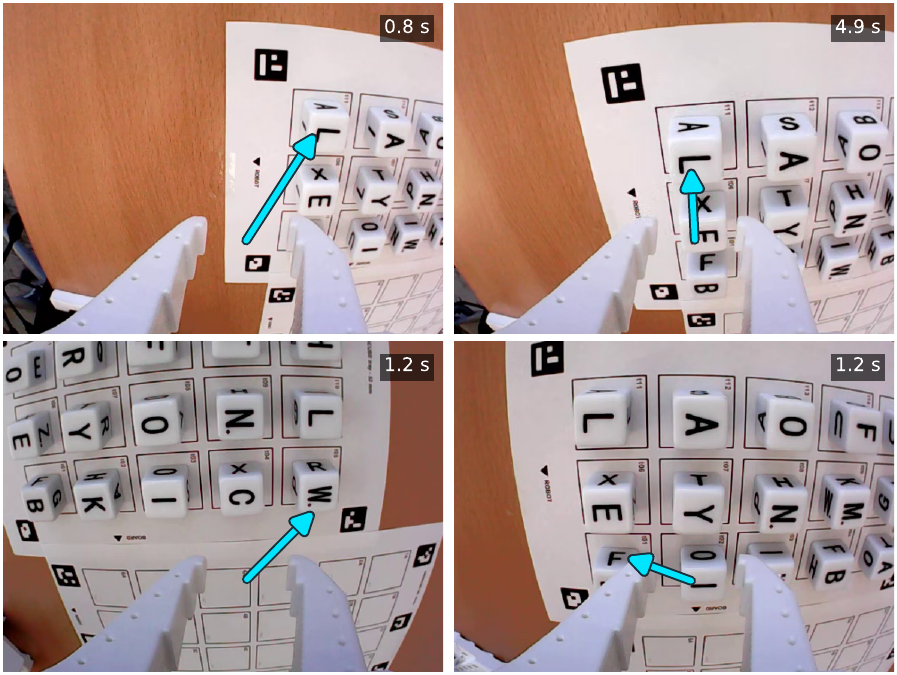}
\caption{Corrective segments for pickup. Arrows illustrate adjustment directions; they are not recorded actions or model inputs.}
\label{fig:corrective-segments-compact}
\end{wrapfigure}

Figure~\ref{fig:bridge1-2} shows that autonomous generation reliability varies across tasks.
GPT-6 Astra succeeds in 8/10 Stack Cups trials, while Fable 5 succeeds in 6/10 Claw Machine trials.
Precise contact remains challenging: on Close a Lid, Fable 5 succeeds in 0/10 trials and Astra in 2/10.
The low success rates on Fold a Cloth and Close a Lid indicate that successful demonstrations alone do not reliably support generation across all four tasks.

\textbf{Improving Precision through Corrective Demonstration Segments.}
Minor spatial misalignments during precise manipulation can lead to missed grasps or failed insertions.
We augment the in-context examples with \textbf{corrective demonstration segments}: short demonstrations focused on moments when execution requires a corrective adjustment.
These examples show how to use visual feedback to correct physical errors, providing guidance beyond successful task executions.
Figure~\ref{fig:corrective-segments-compact} illustrates corrective alignment during cube pickup; Appendix~\ref{app:corrective-segments} reports per-task segment counts and additional examples (Figure~\ref{fig:corrective-segment-examples}).

\begin{figure}[t]
\centering
\includegraphics[width=\linewidth,height=0.72\textheight,keepaspectratio]{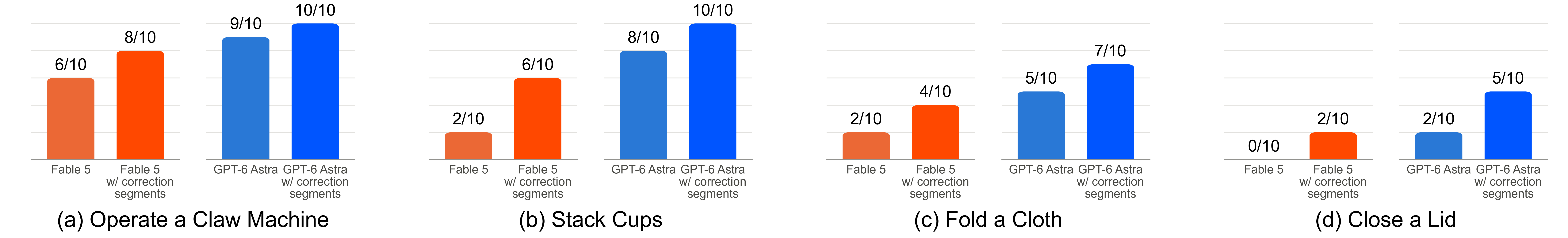}
\caption{Effect of adding corrective demonstration segments to the in-context examples. Success counts compare the baseline and augmented conditions for Fable 5 and GPT-6 Astra across four manipulation tasks, with ten trials per condition.}
\label{fig:bridge1-3}
\end{figure}

Corrective demonstration segments improve observed success across the evaluated tasks (Figure~\ref{fig:bridge1-3}).
On Close a Lid, Astra improves from 2/10 to 5/10 successful trials, while Fable 5 improves from 0/10 to 2/10.
Astra also succeeds in all ten Stack Cups trials.
These results suggest that examples of visual corrections help recover attempts that would otherwise fail, although precise contact remains challenging.

\subsection{Data Synthesis via Autonomous Generation}
\label{sec:generation-efficiency}

We use GPT-6 Astra to collect successful manipulation trajectories for local-policy training.
During synthetic trajectory collection, frontier-controlled episodes can take tens of minutes, with some exceeding 100 minutes. These data-collection runs are distinct from the one-hour capability trials in Section~\ref{sec:autonomous-rollouts}.
We therefore examine whether accumulating successful examples in context reduces generation time and API cost.

\begin{figure}[t]
\centering
\includegraphics[width=\linewidth,height=0.72\textheight,keepaspectratio]{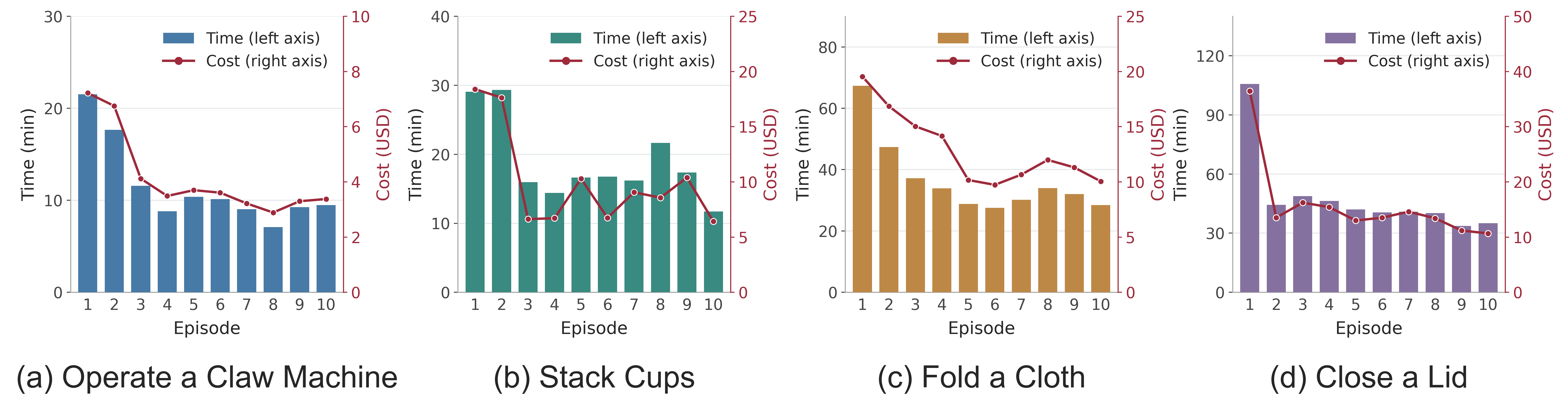}
\caption{Generation efficiency as successful demonstrations accumulate in context. Bars show time in minutes (left axis), and lines show API cost in USD (right axis) for GPT-6 Astra over ten successful episodes per task. Episode indices count successful rollouts; time and API cost from failed attempts are excluded.}
\label{fig:bridge1-4}
\end{figure}

\textbf{Generation Efficiency with Accumulated Examples.}
We track time and API cost across ten successful episodes per task.
Unlike the independent trials with five demonstrations in Section~\ref{sec:autonomous-rollouts}, these generation sessions retain context across episodes.
Useful segments from successful trajectories are added to a separate pool while the original five human demonstrations are retained. The model reuses and adapts these segments when the current situation matches, updating its strategy after each episode. Section~\ref{app:experience-prompt} provides the additional instruction used for this experiment.
Figure~\ref{fig:bridge1-4} reports time and API cost per successful episode; time and API cost from failed attempts are excluded.
These measurements describe successful-rollout efficiency rather than the total cost of collecting the dataset.

Generation time and API cost drop sharply after the first one or two successful episodes (Figure~\ref{fig:bridge1-4}).
On Close a Lid, both decrease by more than 50\% between the first and second episodes; on Stack Cups, API cost falls from approximately \$18 to \$7 between the second and third episodes.
Subsequent episodes generally maintain lower time and cost despite episode-to-episode variation.
These observations suggest that retaining context across episodes can improve generation efficiency.

\subsection{Harnessing Frontier Models for Local Policies}

We incorporate autonomously generated trajectories into local-policy training alongside human demonstrations.
For each task, the training corpus contains 40 successful episodes: 20 human-teleoperated demonstrations and 20 trajectories generated by GPT-6 Astra.
Appendix~\ref{app:human-synthetic-comparison} compares this mixed corpus with 40 human demonstrations per task; both achieve 41/80 policy-alone successes across the four tasks (Figure~\ref{fig:human-synthetic-comparison}).
The autonomous trajectories include corrective movements, exposing the local policy to error-recovery behavior during training.

At deployment, the frontier model receives frames from three camera views, the task description, and the goal condition, and provides high-level instructions to the local policy.
We fine-tune a pretrained OpenPI $\pi_{0.5}$ policy~\citep{black2025pi05} on the mixed demonstration corpus described above to obtain the local controller, which predicts ten-action chunks of absolute joint targets.
Appendix~\ref{app:autonomous-generation} documents the training corpus, policy architecture, and deployment interface, including timing limitations.

To evaluate the integrated harness during deployment, we compare four control configurations: GPT-6 Astra alone, Astra with corrective demonstration segments (CS), the local policy alone (Policy-alone), and the harness with Astra.
We evaluate 20 trials per configuration and task, reporting success over all trials and mean elapsed time over successful trials only. Time includes frontier reasoning, planning, tool calls, API latency, and corrective retries, but excludes hardware-fix intervals (Section~\ref{app:physical-timing}).

\begin{figure}[t]
\centering
\includegraphics[width=\linewidth,height=0.72\textheight,keepaspectratio]{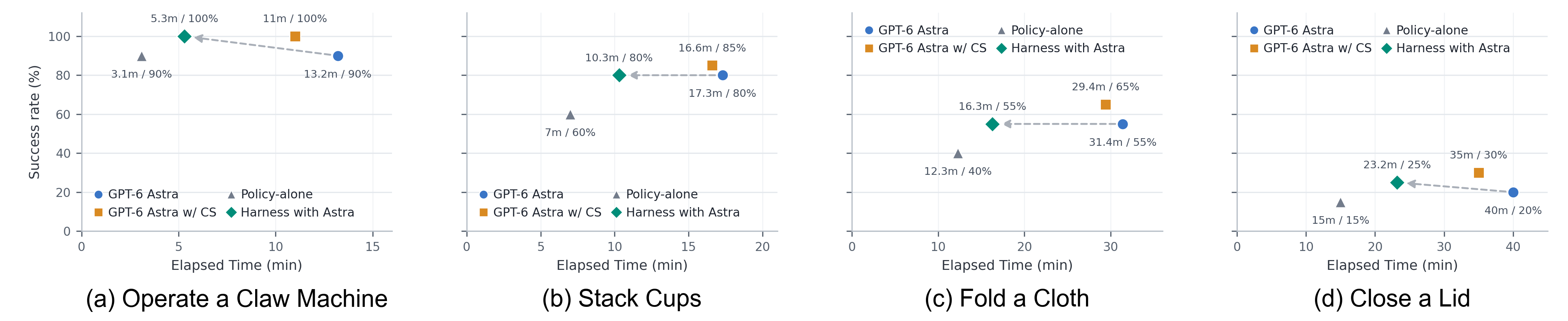}
\caption{Average elapsed task time and task success across four control configurations: GPT-6 Astra alone, Astra with corrective demonstration segments (CS), the local policy alone, and the harness with Astra. Each panel reports one manipulation task, with $n=20$ trials per configuration and task. Time is averaged over successful trials only and includes reasoning, planning, tool calls, and API latency, excluding hardware-fix intervals. Success rate uses all 20 trials.}
\label{fig:bridge1-5}
\end{figure}

Figure~\ref{fig:bridge1-5} compares execution time and success rate across the four control configurations.
Across the four tasks, the harness achieves 65\% success (52/80 trials), compared with 51.25\% (41/80) for the local policy alone.
Relative to GPT-6 Astra alone, mean elapsed time over successful trials is approximately 40--60\% lower.
For example, on Fold a Cloth, time decreases from 31.4 to 16.3 minutes while maintaining the same 55\% success rate.
These results support separating frontier-model guidance from local motor execution to balance task success and execution time, although precise contact remains challenging.
The effectiveness of this separation depends on the local policy's ability to follow the frontier model's instructions, motivating the language-supervision study in Section~\ref{sec:bridge2}.

\section{Bridge 2: Expanding the Policy Language Channel}
\label{sec:bridge2}

Within the harness, the frontier model provides high-level instructions while the local policy executes motor actions.
The effectiveness of this division depends on whether the policy responds to the intended instruction granularity and wording.
We address these requirements through supervision along two axes:
\textbf{1) Temporal resolution}: Task-level or subtask-level descriptions provide little supervision for individual movements.
\textbf{2) Linguistic diversity}: A single templated description per segment does not capture the varied instructions produced by a frontier planner.
We introduce dense language supervision by annotating the same demonstration episodes at three nested granularities---primitive, atomic, and composite---and pairing each level with multiple instruction formats and rephrasings.

\subsection{A Corpus at Three Hierarchical Granularities}

We annotate the same demonstration episodes at three nested granularities: (1) a \textbf{composite} annotation names the \emph{full multi-stage} task; (2) an \textbf{atomic} annotation names one \emph{subtask} within it; and (3) a \textbf{primitive} annotation names a \emph{single-skill segment} within a subtask, such as a reach, a grasp, or a carry.
The three layers describe nested temporal spans within the same episodes.
Table~\ref{tab:annotation-levels}(a) illustrates the hierarchy for \texttt{MakeFruitBowl}, retaining the complete task and subtask sequence and expanding the apple-transfer subtask into primitive instructions.
The other subtasks have their own primitive annotations, omitted from the example for space.
The annotations describe the same underlying trajectories at different temporal resolutions.
This provides a vocabulary for individual movements as well as complete subtasks.

\begin{table}[t]
\caption{Illustrative language supervision for \texttt{MakeFruitBowl}, an example task from RoboCasa 365, with an apple and a banana in the cabinet and a bowl on the counter. (a) The complete task and subtask sequence, with primitives shown only for apple transfer. (b) Four descriptions of that same apple-transfer subtask. Object names and kinematic descriptions must match the demonstrated episode.}
\label{tab:annotation-levels}
\centering
\small
\setlength{\tabcolsep}{3pt}
\renewcommand{\arraystretch}{1.1}
\begin{tabularx}{\linewidth}{@{}p{0.23\linewidth}>{\raggedright\arraybackslash}X@{}}
\toprule
\multicolumn{2}{@{}l}{\textbf{(a) Temporal granularity}} \\
\midrule
Composite\newline(task-level) & Open the cabinet. Place the apple and banana from the cabinet into the bowl. Then close the cabinet. \\
Atomic\newline(subtask-level) & Open the cabinet \(\rightarrow\) \colorbox{appletransfer}{\textbf{transfer the apple into the bowl}} \(\rightarrow\) transfer the banana into the bowl \(\rightarrow\) close the cabinet. \\
\rowcolor{appletransfer}
Primitive\newline(apple transfer only) & Reach for the apple \(\rightarrow\) grasp the apple \(\rightarrow\) lift the apple \(\rightarrow\) carry it over the bowl \(\rightarrow\) lower the apple \(\rightarrow\) release it into the bowl \(\rightarrow\) retract the arm. \\
\midrule
\multicolumn{2}{@{}l}{\textbf{(b) Instruction formats for apple transfer}} \\
\midrule
Arm/gripper kinematics & Grasp the apple, lift it out of the cabinet, carry it over the bowl, and release it into the bowl. \\
Spatial destinations & Move the apple from inside the cabinet into the bowl on the counter. \\
Terse command & Put the apple in the bowl. \\
Natural phrasing & Take the apple out of the cabinet and pop it into the bowl. \\
\bottomrule
\end{tabularx}
\end{table}

RoboCasa 365 provides both atomic and composite annotations; we additionally construct primitive-level annotations.
We derive primitive segments from proprioceptive signals, using gripper opening/closing and end-effector motion to identify physical events, with short boundary buffers and fallback windows.
This yields eight primitive kinds over 27,113 segments, drawn from 4,100 episodes across 91 tasks on RoboCasa 365. For BEHAVIOR-1K, we use the primitive annotations and corresponding segments provided by the benchmark.

\subsection{Dense Multi-aspect Language Supervision}

We adapt the multi-aspect supervision approach from DeMiAn \cite{demian2026} to provide four instruction formats for each segment: arm/gripper kinematics, spatial destinations, a terse command, and natural phrasing.
These formats emphasize how the robot moves, where objects move, minimal wording, and conversational wording, respectively.
Table~\ref{tab:annotation-levels}(b) illustrates all four for the same apple-transfer subtask.
These formats provide complementary descriptions of the same action segment, varying in both wording and detail.
Appendix~\ref{app:corpus-quality} reports corpus construction and physical consistency checks, and Section~\ref{app:annotation-audit} describes VLM screening and human review of object-identity annotations.

\subsection{Experimental Setup}

We evaluate our language-supervision strategies on two simulated benchmarks, RoboCasa 365~\citep{nasiriany2026robocasa365,robocasa365} and BEHAVIOR-1K~\citep{li2023behavior1k}. We choose simulation-based benchmarks that provide long-horizon tasks requiring robust high-level planning and enable controlled comparisons of language inputs. This allows us to vary instruction granularity and phrasing while keeping initial conditions and evaluation criteria consistent.
1) \textbf{RoboCasa 365}: We evaluate ten composite kitchen tasks, scored via strict success rate over 240 paired episodes per seed (24 episodes per task).
2) \textbf{BEHAVIOR-1K}: We evaluate eight repeated rollouts of one instance for each of seven tasks (56 episodes per condition and seed). Due to the extreme difficulty of this benchmark, where complete task successes approach zero, we report mean subgoal progress (mean $q$), the benchmark's predicate-based partial-success score, rather than binary success rates. Here, subgoals refer to benchmark goal predicates, not planner-generated instruction steps.
To ensure fair comparison, all policies are initialized from the same pretrained checkpoint and trained for a matched budget of 120,000 steps, with training and evaluation seeds 42 and 142 for the main language-supervision comparison.
Table~\ref{tab:supervision-recipes} in Appendix~\ref{app:evaluation-protocol} defines the prompt content for each condition.
Benchmark tasks and training configuration details are provided in Appendix~\ref{app:evaluation-protocol}.

\textbf{Instructor conditions.} We evaluate each language supervision recipe under two instructor conditions. Under the \textbf{oracle} instructor, the subgoal plan is authored and instruction switching follows simulator-state milestones, providing a privileged reference for policy performance.
Under \textbf{Fable 5}, the frontier model serves as both planner and instructor: it determines the subgoal sequence and provides the corresponding language instructions for the local policy.
The full sequence is generated once at the start of the episode from the composite goal text and reset-frame camera views, without access to simulator state.
For each evaluation episode, the sequence is generated once and reused across all policy variants, keeping instruction content fixed across supervision conditions.
We separate progress monitoring from high-level planning~\citep{yi2026critic}.
A local open-weights VLM, Qwen3.6-27B, monitors execution from camera observations and decides when the current subgoal is complete.
This makes routine progress checks more efficient by avoiding frequent calls to the slower frontier model.
All policy variants share this progress monitor.
When completion is confirmed, the harness passes the next instruction from the original plan to the local policy.
Appendix~\ref{app:frontier-instructor} details this protocol and its connection to Probe 2.

\Needspace{23\baselineskip}
\subsection{Results}
\label{sec:language-results}

\begin{wraptable}{r}{0.62\textwidth}
\caption{Effect of language supervision on long-horizon task performance under oracle and frontier-model instructors. Values average seeds 42 and 142: success rate on RoboCasa 365 and mean subgoal progress on BEHAVIOR-1K. Table~\ref{tab:language-results-seeds} gives both seed results.}
\label{tab:language-results}
\centering
\footnotesize
\setlength{\tabcolsep}{3pt}
\renewcommand{\arraystretch}{1.05}
\begin{tabularx}{\linewidth}{@{}>{\raggedright\arraybackslash}Xrrrr@{}}
\toprule
 & \multicolumn{2}{c}{\textbf{RoboCasa 365}} & \multicolumn{2}{c}{\textbf{BEHAVIOR-1K}} \\
\cmidrule(lr){2-3}\cmidrule(lr){4-5}
Language Supervision & Oracle & Fable 5 & Oracle & Fable 5 \\
\midrule
Task-level (fixed) & \multicolumn{2}{c}{0.169} & \multicolumn{2}{c}{0.076} \\
Subtask-level & 0.406 & 0.381 & 0.143 & 0.084 \\
Task + Subtask & 0.417 & 0.402 & 0.154 & 0.091 \\
\midrule
Hierarchical only & 0.456 & 0.440 & 0.143 & 0.084 \\
Multi-aspect only & 0.425 & 0.406 & 0.160 & 0.103 \\
Hierarchical + Multi-aspect & \textbf{0.517} & \textbf{0.485} & \textbf{0.174} & \textbf{0.143} \\
\bottomrule
\end{tabularx}
\end{wraptable}

Table~\ref{tab:language-results} shows that subtask instructions improve performance over fixed task-level instructions. Adding task-goal context improves both benchmarks. Hierarchical supervision further improves RoboCasa success over Task + Subtask, but not BEHAVIOR-1K progress; multi-aspect supervision alone improves both benchmarks. The combined recipe performs best under both instructors, reaching oracle scores of 0.517 and 0.174 on RoboCasa and BEHAVIOR-1K, respectively. Under Fable 5, it improves the corresponding scores from 0.381 to 0.485 and from 0.084 to 0.143 over subtask-level supervision.

\subsubsection{Probe 1: Multi-Granularity Instruction Following}

We first probe whether hierarchical supervision improves the local policy's ability to follow fine-grained movement commands.
We evaluate single-seed subtask-level and hierarchical-only checkpoints from the main evaluation on six primitive movement commands across 12 RoboCasa scenes, without additional training.
We measure instruction compliance by checking whether the final displacement along the commanded axis meets a predefined threshold.
Appendix~\ref{app:primitive-compliance} provides the protocol and results (Table~\ref{tab:primitive-compliance-full}). 
Hierarchical supervision improves compliance across all six commands, including ``move the arm down'' (0.111 to 0.778) and ``move the arm back'' (0.028 to 0.444).
These results suggest that hierarchical supervision broadens the policy's language channel to better support primitive movement commands alongside subtask-level instructions.
We note that this probe measures directional movement, not successful manipulation or online correction during complete tasks.

\Needspace{5\baselineskip}
\subsubsection{Probe 2: Robustness to Instruction Diversity}

A robust local policy must understand the semantic intent of an instruction regardless of the specific syntax or vocabulary used by the instructor.
We test two forms of instruction variation: changing the source planner among Fable 5, Opus 5, and Gemini-3.1-Pro, and rephrasing the resulting instructions with Fable 5.
In the rephrased condition, we prompt Fable 5 to preserve the original task goal while changing how the instructions are expressed.
For example, ``Open up the cabinet'' becomes ``Grasp the cabinet door handles and swing the doors open.''
This condition tests goal-preserving variation in wording and descriptive detail; Fable 5 serves as the rephrasing model for the instructions from all three source planners.
We first evaluate the quality of the complete high-level plans generated once at the start of each episode by these three frontier models, using the composite goal text and reset-frame camera views.
We compare their predicted plans across 30 scenes against the ground-truth oracle plans.
Within each planner and phrasing condition, the generated plan for an evaluation episode is reused across all policy variants; observation-based completion monitoring and instruction switching follow the shared protocol in Appendix~\ref{app:frontier-instructor}.

Table~\ref{tab:planner-quality} in Appendix~\ref{app:frontier-instructor} reports decomposition step-count match and semantic similarity, showing that all three models generate plans highly similar to the oracle reference.
These text-based metrics provide a check on plan quality, but do not establish that the plans are functionally equivalent. Execution differences across planners can therefore reflect both wording and differences in plan content or ordering.
Table~\ref{tab:planner-robustness} reports single-seed end-to-end execution success across the three planners and their rephrased instruction conditions.

\begin{table}[t]
\caption{Task success rates on RoboCasa 365 by source planner (seed 42). Rephrased instructions are generated by Fable 5, preserving the task goal while changing the wording. Parentheses report changes between the displayed original and rephrased rates.}
\label{tab:planner-robustness}
\centering
\small
\setlength{\tabcolsep}{3pt}
\renewcommand{\arraystretch}{1.15}
\begin{tabular*}{\linewidth}{@{\extracolsep{\fill}}l|cc|cc|cc@{}}
    \toprule
    Language Supervision & \multicolumn{2}{c|}{Opus 5} & \multicolumn{2}{c|}{Gemini-3.1-Pro} & \multicolumn{2}{c}{Fable 5} \\
    & Original & Rephrased & Original & Rephrased & Original & Rephrased \\
    \midrule
    Subtask-level & .346 & .333 (-.013) & .325 & .329 (+.004) & .392 & .350 (-.042) \\
    Hierarchical only & .413 & .408 (-.005) & .400 & .388 (-.012) & .446 & .400 (-.046) \\
    Multi-aspect only & .371 & .363 (-.008) & .350 & .350 (.000) & .417 & .408 (-.009) \\
    Hierarchical + Multi-aspect & \textbf{.488} & \textbf{.471 (-.017)} & \textbf{.467} & \textbf{.454 (-.013)} & \textbf{.492} & \textbf{.483 (-.009)} \\
    \bottomrule
    \end{tabular*}
\end{table}

For the subtask-level baseline, changing the planner from Fable 5 to Gemini-3.1-Pro reduces success from 0.392 to 0.325, while using the rephrased versions of Fable 5's instructions reduces success from 0.392 to 0.350.
Multi-aspect supervision reduces sensitivity to rephrasing across all three planners, with absolute changes in success of at most 0.009, compared with up to 0.042 for the subtask-level baseline.
The combined recipe achieves the highest success in all six conditions (0.454--0.492); even its lowest success exceeds the baseline's highest (0.392).
Together, these results suggest that diverse language supervision makes the policy less sensitive to how instructions are expressed, while combining it with hierarchical supervision sustains higher task success across the tested instructors and instruction phrasings.

\section{Bringing Both Bridges Together}
\label{sec:crossword}
\begin{figure}[t]
\centering
\includegraphics[width=\linewidth]{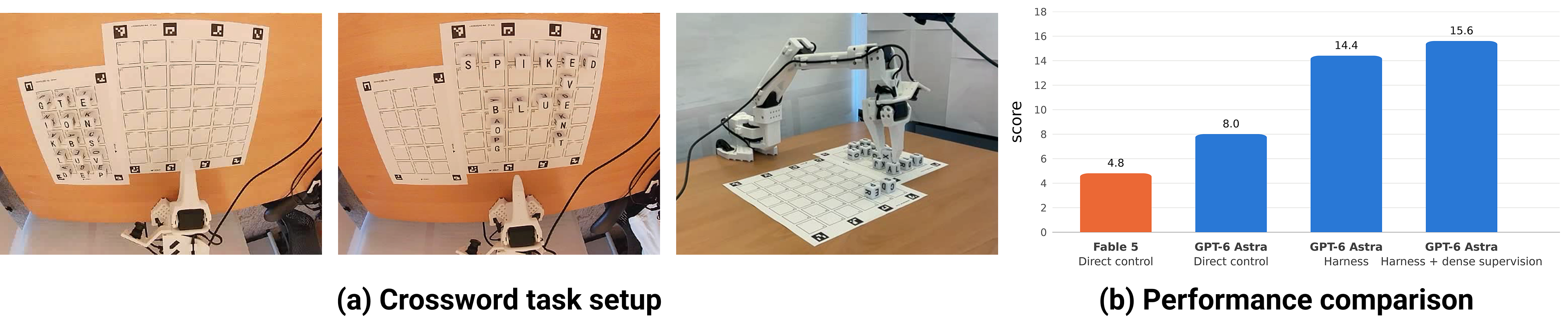}
\caption{The crossword task. (a) Example board states and the robot workspace. (b) Mean total score after one hour across five games for four control configurations. Per-game scores are reported in Table~\ref{tab:crossword-per-game}.}
\label{fig:crossword-results}
\end{figure}

We evaluate both bridges together on the crossword task (Figure~\ref{fig:crossword-results}(a)), which requires forming valid words and placing alphabet cubes on a shared board. Each placement changes the board and remaining letter pool. In our qualitative observations, both frontier models proposed valid words, while physical execution was the main bottleneck.

\textbf{Setup.} The workspace contains an initially empty $6\times6$ grid and a randomized pool of 15 alphabet cubes.
We use GPT-6 Astra to autonomously generate a demonstration corpus for the crossword task and annotate the trajectories with dense, multi-aspect language supervision.
To assess the contribution of dense language supervision, we compare two harness configurations: one with a local policy trained on subtask-level instructions and the other with hierarchical and multi-aspect supervision.
Both use the same demonstrations, policy architecture, training budget, and frontier controller.
During harness deployment, the frontier model reads camera views at times of its choosing and issues language instructions with execution durations; the local policy generates motor actions.
Subsequent observations support corrective instructions after failed pickups and revision of planned placements.
Unlike the fixed-plan evaluation in Section~\ref{sec:bridge2}, instructions can be revised during execution.
Appendix~\ref{app:crossword-results} provides the policy training configuration and a separate 20-trial pick-and-place evaluation (Table~\ref{tab:crossword-pick-place}).
We evaluate performance using two reward components: $+2$ points for each cube successfully placed at its instructed grid coordinate and $+2$ points for each valid word formed.

\textbf{Results.} We report the mean total score over five games, each using a different alphabet set and a one-hour execution budget, for four control configurations: Fable 5 alone, GPT-6 Astra alone, the GPT-6 Astra harness, and the GPT-6 Astra harness with dense language supervision.
Appendix~\ref{app:crossword-results} provides per-game results and an additional comparison prohibiting cell IDs in instructions (Table~\ref{tab:crossword-per-game}).
We omit policy-alone execution because the task requires high-level word planning.
As shown in Figure~\ref{fig:crossword-results}(b), both harness configurations achieve higher scores than either frontier model alone within the fixed time budget.
The dense-supervision variant further improves the mean score.
When board cell IDs are prohibited, the frontier model must issue instructions such as ``put the cube to the left of the B cube'' instead of ``put the cube on 1A.''
Under this restriction, dense supervision increases the mean score from 2.8 to 5.6 across five games (Appendix~\ref{app:crossword-results}), although both policies perform worse than with cell IDs.
Together with the multi-granularity and instruction-diversity probes in Section~\ref{sec:language-results}, these results suggest that broadening the policy's language channel supports execution across fine-grained movement commands, rephrased subgoals, and spatial descriptions of target locations.

\Needspace{18\baselineskip}
\begin{wrapfigure}{r}{0.56\textwidth}
\centering
\includegraphics[width=\linewidth]{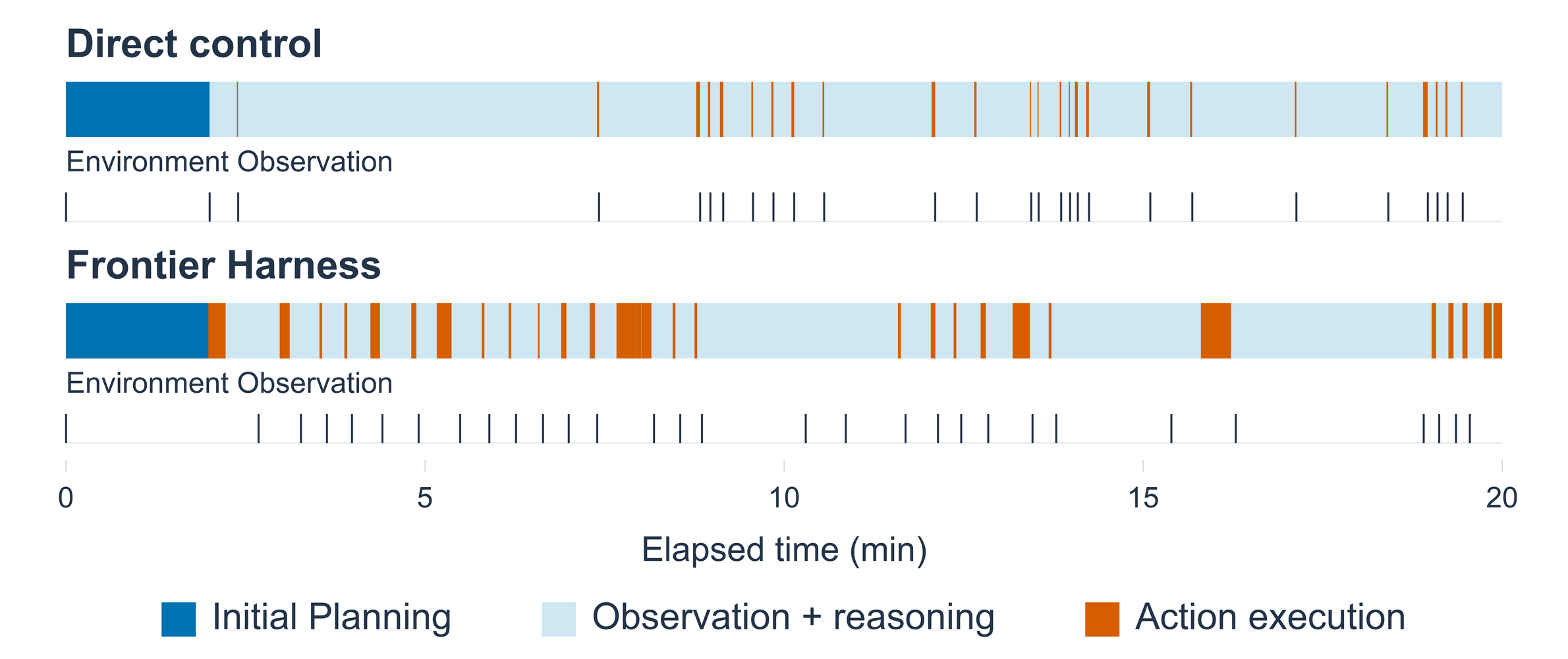}
\caption{First 20 minutes of crossword execution under direct control and the frontier harness, both using GPT-6 Astra.
Colored intervals show the frontier model's initial planning and subsequent observation and reasoning, as well as robot action execution.
Ticks mark model-requested camera observations.
Each timeline represents one run per configuration.
Full 60-minute timelines appear in Figure~\ref{fig:crossword-activity-full}.}
\label{fig:crossword-activity}
\end{wrapfigure}
\textbf{Execution timing.} The harness delegates motor execution to a low-latency local policy, but the overall system still pauses for frontier reasoning between execution segments.
Figure~\ref{fig:crossword-activity} shows the first 20 minutes of the direct-control and frontier-harness runs.
Over the full one-hour runs, cumulative action execution increases from 2 min 40 s under direct control to 9 min 8.55 s with the harness.
This corresponds to approximately 3.4 times as much time spent executing robot actions within the same budget, illustrating the benefit of delegating motor execution to the local policy.
Appendix~\ref{app:crossword-results} provides timing details.

\section{Conclusion}
We study two complementary bridges between frontier reasoning and robot execution: autonomous demonstration generation to supplement local-policy training, and dense language supervision to improve instruction following.
We introduce corrective demonstration segments to improve generation reliability.
Generation time and cost decrease across the observed successful episodes as examples accumulate.
Combining hierarchical and multi-aspect language supervision improves instruction following on long-horizon simulation tasks.
Evaluation on the crossword task brings both bridges together, with the harness achieving higher scores than direct frontier control within a fixed time budget.
Delegating motor execution to a local policy mitigates the impact of frontier-inference latency, allowing more of the available time to be spent on physical execution.
Extending this approach to broader physical tasks and reducing dependence on human demonstrations remain directions for future work.
Together, these results suggest that effective delegation benefits from both capable local execution and a language interface that supports the frontier model's guidance.

\newpage

\bibliography{references}
\bibliographystyle{iclr2027_conference}

\appendix
\section{Limitations and Future Work}
\label{app:limitations}
\textbf{Experimental scope.} The main language-supervision comparison uses two training/evaluation seed pairs, providing limited replication without separating training and evaluation variability. Evaluation tasks are represented in training: RoboCasa 365 tests held-out layouts, styles, and sampled assets, whereas BEHAVIOR-1K uses one held-out initial configuration per task within existing scenes. Physical evaluations cover four manipulation tasks and five crossword games; broader task coverage and additional independent runs are needed to assess generality.

\textbf{Autonomous demonstration collection.} Policies trained on 20 human and 20 generated demonstrations per task attain the same aggregate success as those trained on 40 human demonstrations (Appendix~\ref{app:human-synthetic-comparison}). This result does not establish statistical equivalence or isolate the incremental benefit over 20 human demonstrations alone, which we do not evaluate. We are collecting additional autonomous demonstrations to evaluate policies trained entirely on generated trajectories.

\textbf{Timing and deployment.} Frontier reasoning introduces pauses between execution segments due to API latency, so low-latency local control does not make the complete system real-time. Generation-efficiency measurements exclude failed attempts and describe observational trends across successful episodes, rather than total collection cost or a causal effect of accumulating examples. Physical deployment times are conditional on success and exclude hardware-fix intervals. More responsive coordination between frontier reasoning and local execution is an important direction for future work.

\section{Interfaces for Frontier-Guided Robot Control}
\label{app:control-taxonomy}
\begin{figure}[htbp]
\centering
\includegraphics[width=\linewidth]{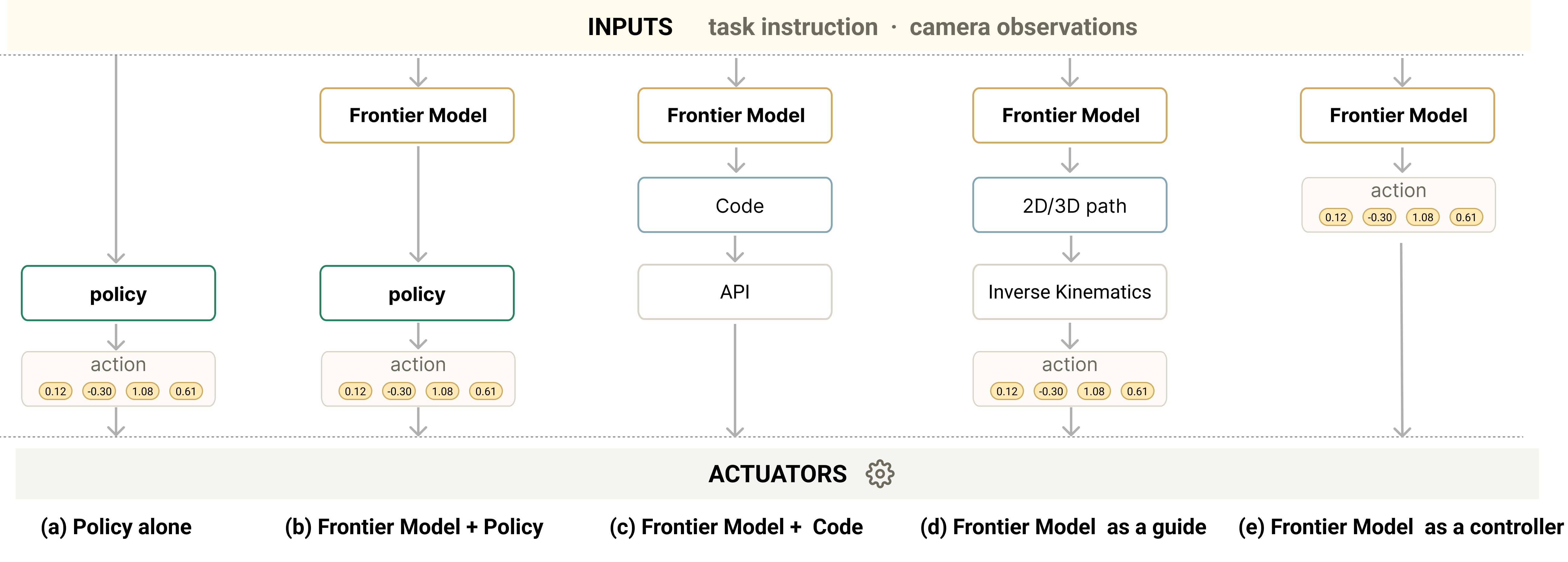}
\caption{Five interfaces for robot control, distinguished by the frontier model's output and the components that translate it into physical actions.}
\label{fig:route-taxonomy}
\end{figure}

Figure~\ref{fig:route-taxonomy} distinguishes five routes to robot control: a policy without a frontier model (a), frontier-generated conditioning for a trained policy (b), executable programs over a fixed API (c), geometric guides (d), and direct joint commands (e). The distinction is what the frontier model emits and how that output reaches the actuators. Geometric guides may be followed by a policy or converted to joint commands through inverse kinematics, so type (d) is defined by the guide rather than its receiver.

\section{Autonomous Demonstration Generation}
\label{app:autonomous-generation}
\label{app:corrective-segments}
\subsection{Autonomous Rollout Prompt}
\label{app:rollout-prompt}

The following prompt is used for autonomous rollouts in Section~\ref{sec:autonomous-rollouts}. Angle-bracketed fields denote the task instruction, task-specific success condition, and data paths.

\begin{quote}
\small
Complete ``\texttt{<TASK\_INSTRUCTION>}'' with an SO-101 robot, using \textbf{five successful demonstrations} from \texttt{<DEMONSTRATION\_DATASET>} as in-context references.

\begin{enumerate}
\item \textbf{Learn and act.} Review all five demonstrations and observe the current scene. Infer a suitable strategy, execute with visual feedback, and adapt to the robot's actual response. Preserve calibrated hardware limits and stop when safe execution cannot be established.
\item \textbf{Verify success.} Confirm that \texttt{<SUCCESS\_CONDITION>} is satisfied.
\item \textbf{Record trajectories.} Save every attempt in \texttt{<OUTPUT\_DIRECTORY>}, including measured states, executed actions, camera observations, timestamps, source references and outcome. Build the synthetic corpus from verified successes; label failures separately.
\item \textbf{Report results.} Report the outcome, segment reuse, \textbf{episode-only estimated API cost}, \textbf{episode elapsed time}, and \textbf{robot command-execution time} separately. Exclude preparation and postprocessing; state measurement limitations. Create a review video for successful episodes, omitting camera-free intervals.
\end{enumerate}
\end{quote}

\subsection{Accumulated-Experience Prompt Extension}
\label{app:experience-prompt}

For the generation-efficiency experiment in Section~\ref{sec:generation-efficiency}, the following instruction is added to the rollout prompt:

\begin{quote}
\small
\textbf{Improve through experience.} Add useful successful segments to \texttt{<SEGMENT\_POOL>}, separate from the original demonstrations. Reuse and adapt them when the current situation matches. Update the strategy after each episode to improve reliability, time and cost.
\end{quote}

The original five demonstrations and the growing segment pool are maintained separately. The model can reuse relevant successful segments in later episodes. The reported generation-efficiency measurements exclude preparation, postprocessing, and failed attempts; they describe successful episodes rather than the total cost of corpus construction. Episode elapsed time and robot command-execution time are recorded separately.

\subsection{Corrective Demonstration Segments}

Corrective demonstration segments are short examples of adjustments made when execution encounters physical errors. Table~\ref{tab:corrective-segment-counts} reports the number of segments used for each task. These counts refer to corrective segments, rather than complete successful trajectories or evaluation trials.

\begin{table}[htbp]
\centering
\small
\caption{Number of corrective demonstration segments used per task.}
\label{tab:corrective-segment-counts}
\begin{tabular}{lr}
\toprule
Task & Corrective segments \\
\midrule
Operate a Claw Machine & 8 \\
Stack Cups & 8 \\
Fold a Cloth & 16 \\
Close a Lid & 16 \\
Physical Crossword Puzzle & 32 \\
\bottomrule
\end{tabular}
\end{table}

\begin{figure}[htbp]
\centering
\includegraphics[width=\linewidth]{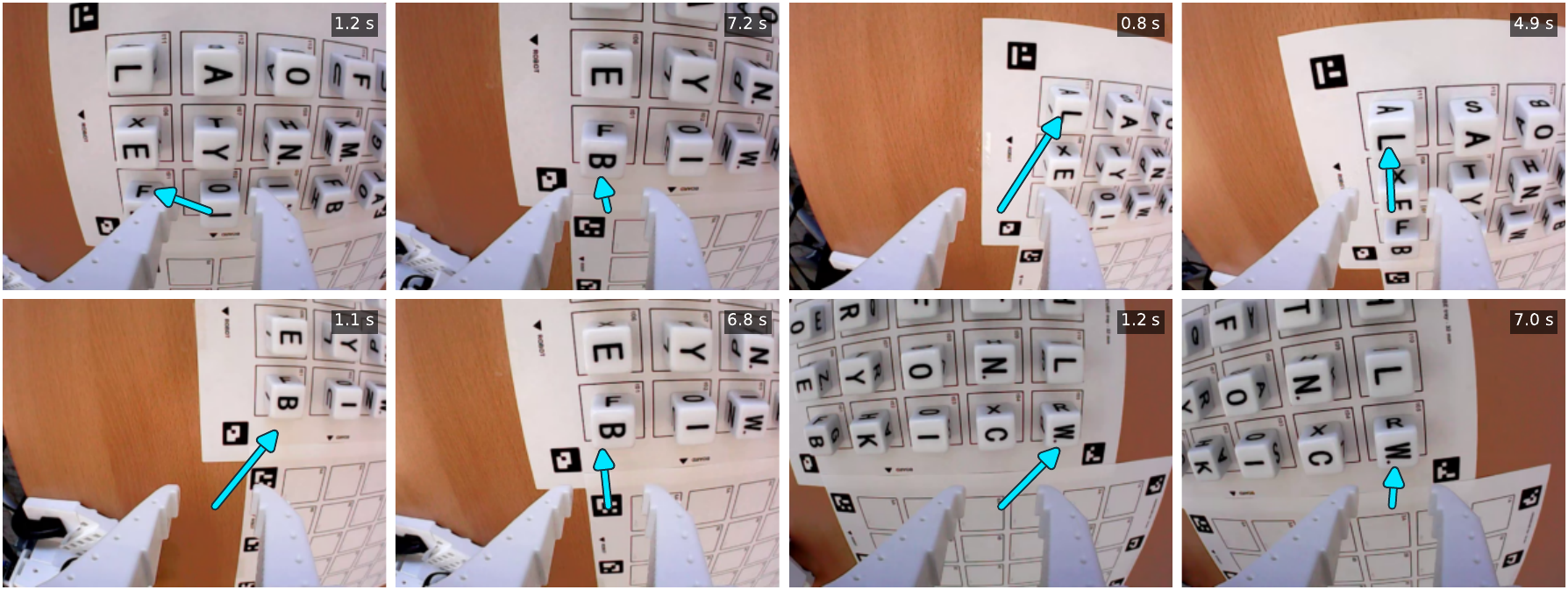}
\caption{Illustrative corrective demonstrations for crossword cube pickup. Each adjacent pair shows the initial offset (left) and subsequent approach (right) in a recorded wrist-camera sequence. The top row shows B and L cube pickups; the bottom row shows a second B cube pickup and a W cube pickup. Cyan arrows indicate illustrative alignment directions toward the target, manually added for visualization; they are not recorded action vectors or overlays supplied to the model. Timestamps indicate elapsed time within each episode.}
\label{fig:corrective-segment-examples}
\end{figure}

Figure~\ref{fig:corrective-segment-examples} illustrates alignment toward target cubes from different initial gripper offsets in four crossword pickup demonstrations.

\subsection{Physical Policy and Deployment Interface}
\label{app:physical-deployment}

We fine-tune a pretrained OpenPI $\pi_{0.5}$ vision-language-action policy on the mixed corpus of human and autonomously generated demonstrations to obtain the physical controller. The policy uses a PaliGemma backbone (\texttt{gemma\_2b}) and a flow-matching action expert (\texttt{gemma\_300m}). It controls an SO-101 arm using overhead and wrist RGB images, six measured joint/gripper state values, and a natural-language instruction. The deployment interface uses $224\times224$ RGB images and duplicates the overhead frame into the policy's two external-camera input slots. This policy input format is distinct from the frontier model's camera access.

Policy inference runs remotely over WebSocket. Each prediction provides ten absolute joint-target actions; the robot uses six action channels, corresponding to five arm joints and the gripper. The intended action rate is 30 Hz, so ten actions represent approximately 333 ms of motion. The documented scheduling strategy issues actions every 33.3 ms and replaces the unexecuted remainder of a chunk when a new prediction arrives.

The 30 Hz rate is a target rather than a consistently achieved measurement. Early diagnostics reported 3.38 end-to-end cycles per second, while later bench measurements reported control rates of 18.5--19.9 Hz. These are different timing measurements from diagnostic runs and do not establish a uniform achieved rate for the task evaluations. Camera acquisition, preprocessing, communication, and inference all contribute to the realized timing.

The instruction is supplied in the \texttt{prompt} field of every inference request. Updating that field can condition the next action chunk without retraining or reloading the policy. The interface alone does not establish a tested mid-execution interruption protocol or switching latency; the documented offline comparison of policy predictions under different instructions does not demonstrate successful physical task switching.

\subsection{Elapsed-Time Accounting and Limitations}
\label{app:physical-timing}

For the physical deployment comparison, elapsed time is averaged over successful trials only; failed and timed-out trials contribute to success rates but not to the time average. Elapsed time includes frontier reasoning and planning, camera observation and verification, tool calls, API and communication latency, physical action execution, and task-level corrective retries. Hardware-fix intervals are excluded. Consequently, these means measure time conditional on success rather than expected time per attempt or total time to obtain a success.

Even with the same prompt, the frontier agent can choose different sequences of observation, verification, and preparatory actions across trials. Service and communication issues introduce additional variability. The prompt does not enforce an identical operational sequence, and excluding hardware-fix intervals limits what the reported times establish about total deployment effort. We do not attribute this variability to unobserved internal model mechanisms.
The horizontal coordinates in Figure~\ref{fig:bridge1-5} report elapsed task time averaged over successful trials. The crossword activity timeline in Figure~\ref{fig:crossword-activity-full} instead illustrates one run per configuration and is not an average across games.

\subsection{Human-Only and Mixed Demonstration Corpora}
\label{app:human-synthetic-comparison}

\begin{figure}[htbp]
\centering
\includegraphics[width=\linewidth]{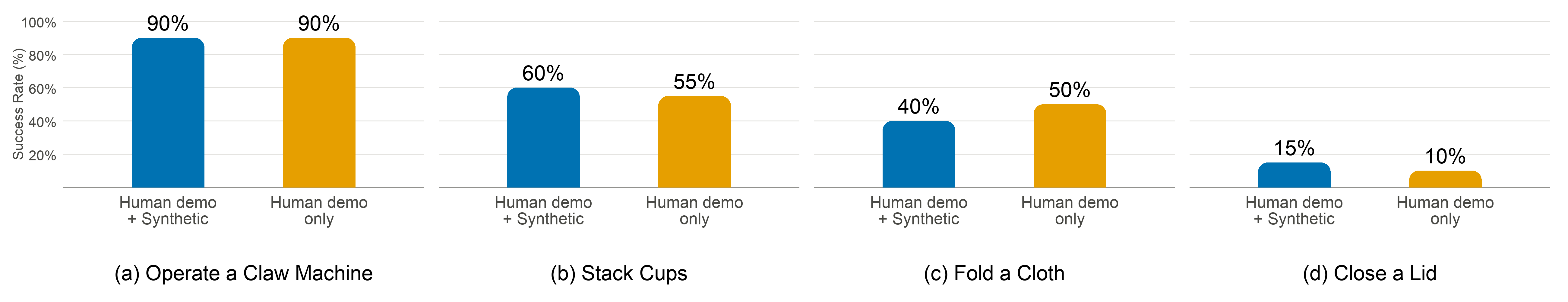}
\caption{Policy success rates with mixed and human-only training corpora. ``Human demo + Synthetic'' uses 20 human and 20 autonomously generated demonstrations per task; ``Human demo only'' uses 40 human demonstrations per task. Each bar reports success over 20 evaluation trials. Both conditions achieve 41/80 successes across the four tasks.}
\label{fig:human-synthetic-comparison}
\end{figure}

We compare policy success after training on 40 human demonstrations per task with training on a mixed corpus of 20 human and 20 autonomously generated demonstrations per task (Figure~\ref{fig:human-synthetic-comparison}). Each condition is evaluated over 20 trials per task. The mixed and human-only corpora yield success rates of 90\% versus 90\% on Operate a Claw Machine, 60\% versus 55\% on Stack Cups, 40\% versus 50\% on Fold a Cloth, and 15\% versus 10\% on Close a Lid. Both conditions achieve 41 successes across 80 trials (51.25\%).

At the same total demonstration count, the mixed corpus uses half as many human demonstrations and attains the same aggregate success in this evaluation, with task-specific differences. This result supports the feasibility of using generated trajectories in place of part of the human training corpus; it does not establish statistical equivalence or consistent improvement across tasks. We do not evaluate a 20-human-only condition, so this comparison does not isolate the incremental benefit of adding generated trajectories to those 20 human demonstrations.

\section{Corpus Construction and Annotation Audit}
\label{app:corpus-quality}
\subsection{Corpus Construction and Physical Validation}
\label{app:construction}

\textbf{Annotation sources and segmentation.} Both RoboCasa 365 and BEHAVIOR-1K provide atomic and composite annotations. BEHAVIOR-1K also provides primitive annotations and their corresponding segments, which we use directly. The primitive segmentation procedure described below applies to RoboCasa 365. For RoboCasa, we additionally construct a primitive corpus containing 27,113 segments from 4,100 episodes across 91 tasks. This is the corpus used for the reported policy training. Segmentation uses recorded action and state streams rather than video or language: gripper opening and closing identify grasp and release, smoothed vertical and horizontal end-effector motion distinguishes lift, lower, and transport, and direction reversals identify shaking. The median segment length is 18 frames, with an interquartile range of 13--35 frames. Table~\ref{tab:primitive-corpus-composition} gives the composition.
\textbf{Segmentation parameters.} We smooth the end-effector trajectory with a five-frame moving average. Motion below 1.5 mm per frame is treated as stationary; otherwise a vertical-to-total motion ratio of at least 0.5 assigns an upward or downward label, with remaining motion assigned to transport. Carry segments (Lift, Lower, and Transport) require at least eight frames, net displacement of at least 2 cm, and a dominant-axis displacement ratio of at least 0.70. A finger gap below 7 cm identifies candidate held spans. Carry boundaries are trimmed to exclude grasp/release intervals and to limit within-segment finger-gap variation to 3 mm.

Grasp and release windows are identified from gripper opening and closing, using a 1 mm per-frame change threshold and a 4 mm net-change check. When that check fails, the implementation falls back to a three-frame window on either side of the event. Windows include two lead-in frames, with trailing buffers of four frames for grasp and eight for release. Reach and retract are bounded by end-effector motion and capped at 80 and 60 frames, respectively; retract begins after release. Shaking merges at least two carry segments when the dominant-axis motion reverses at least three times, the path length is at least 10 cm, net displacement is at most 15 cm, and the net-to-path ratio is at most 0.45.

\begin{table}[htbp]
\centering
\small
\caption{Composition of the RoboCasa primitive annotation corpus.}
\label{tab:primitive-corpus-composition}
\begin{tabularx}{\linewidth}{lrX}
\toprule
Primitive & Segments & Physical interpretation \\
\midrule
Transport & 6,256 & Held object translates approximately horizontally. \\
Grasp & 4,086 & Gripper closes on an object. \\
Reach & 4,066 & Empty arm approaches an object with the gripper open. \\
Release & 3,408 & Gripper opens to release an object. \\
Lower & 3,356 & Held object descends. \\
Retract & 3,057 & Empty arm withdraws after release. \\
Lift & 2,859 & Held object rises. \\
Shake & 25 & Held object oscillates without sustained transport. \\
\midrule
Total & 27,113 & \\
\bottomrule
\end{tabularx}
\end{table}

\textbf{Object identity.} Object labels are derived from recorded simulator state rather than from nouns in task descriptions. Table~\ref{tab:object-identity-provenance} summarizes how these object labels are obtained. For phases with little object displacement, identity can be inherited from transport within the same grasp-to-release span. A consistency check against a directional inheritance rule found one disagreement in 10,314 cases across 222 multi-pick episodes. Unresolved cases retain the decomposition's object name and are explicitly marked; these are concentrated in fixture manipulation, where the relevant part lacks a free joint. Asset mesh categories provide 112 distinct object names.

\begin{table}[htbp]
\centering
\small
\caption{Sources of primitive-segment object labels. Simulator-derived coverage includes inherited identities and is not an annotation-accuracy estimate.}
\label{tab:object-identity-provenance}
\begin{tabularx}{\linewidth}{Xrr}
\toprule
Source & Segments & Share \\
\midrule
Object motion within the segment & 14,935 & 55.1\% \\
Inheritance within the grasp-to-release span & 10,236 & 37.8\% \\
Expanded temporal window & 1,147 & 4.2\% \\
Unresolved; decomposition label retained & 795 & 2.9\% \\
\midrule
Simulator-derived total & 26,318 & 97.1\% \\
\bottomrule
\end{tabularx}
\end{table}

\textbf{Language generation.} We use Opus 5 for annotation generation. The language supervision uses the four formats described in Section~\ref{sec:bridge2}: arm/gripper kinematics, spatial destinations, terse commands, and natural phrasing, with additional rephrasings. Captions are conditioned on the segment's primitive kind and object identity and retained only after validation.

\textbf{Physical consistency.} Table~\ref{tab:primitive-physical-validation} reports checks against the motion signals associated with each primitive. Among 230 sampled transport segments, 96.5\% contain a single moving object. In eight inspected cases with multiple moving objects, the annotation agrees with the object rigidly held by the gripper. The visual audit is described in Section~\ref{app:annotation-audit}.

\begin{table}[htbp]
\centering
\small
\caption{Physical consistency checks for primitive annotations. These checks measure agreement with motion signals, not downstream task success.}
\label{tab:primitive-physical-validation}
\begin{tabularx}{\linewidth}{Xl}
\toprule
Check & Result \\
\midrule
Transport with more than 6 mm of gripper-finger motion & 0.0\%; median 0.7 mm \\
Grasp containing gripper closing & 96.9\% \\
Release containing gripper opening & 93.3\% \\
Retract starting with gripper at least 80\% open & 74.4\%; median openness 0.87 \\
Retract overlapping release & 0 / 3,057 \\
\bottomrule
\end{tabularx}
\end{table}

\textbf{Limitations.} Object identity inherited across phases is inferred rather than directly observed within each segment, and 2.9\% of identities remain unresolved by simulator-based checks. Shaking is rare in the corpus and its translation-reversal detector can miss tilting motions. Visual inspection covers only a sample. These measurements characterize annotation consistency; they do not isolate the causal effect of annotation quality on task success.

\subsection{Annotation Audit}
\label{app:annotation-audit}

\textbf{VLM screening.} We use Opus 5 to inspect manipulated-object identity from RoboCasa's default agent-left, agent-right, and wrist-camera views. The model receives segment imagery without the object label, task name, or annotation slots, and its predicted identity is compared with the simulator-derived annotation. The sample contains 207 segments across 61 tasks, selected round-robin across tasks within each primitive kind: 26 segments for each of seven kinds and all 25 available shake segments. This screening identifies cases for inspection rather than estimating annotation accuracy: the VLM can itself misidentify objects.
Each segment is represented by 12 frames.

\textbf{Human review.} The authors manually reviewed 50 randomly sampled segments and agreed with the annotated object identity in 48 cases (96\%). This result describes agreement within the reviewed sample, rather than a comprehensive assessment of annotation accuracy across the corpus. Physical consistency checks for primitive labels are reported in Section~\ref{app:construction}.

\section{Training and Evaluation Details}
\label{app:evaluation-protocol}
\textbf{Training/evaluation split.} Evaluation tasks are included in the training task sets for both benchmarks. For RoboCasa 365, training uses pretrain demonstrations, whereas evaluation uses the target environment split with held-out kitchen layouts, visual styles, and sampled object assets. For BEHAVIOR-1K, the seven evaluation tasks belong to the ten-task training subset; evaluation uses a held-out public-test initial configuration, instance 301, for each task. This tests new object and robot initial configurations within the existing scene setup. The seven-task results use repeated rollouts of this single configuration per task and do not establish generalization across unseen houses or layouts.

\textbf{RoboCasa evaluation.} The evaluation set contains ten composite tasks selected to test multi-step execution and instruction switching across increasing task depth (Table~\ref{tab:robocasa-task-depth}). It includes cabinet and drawer articulation, repeated object transfers, and rinsing, with three tasks containing two subgoals, three containing three subgoals, and four containing four subgoals. Training demonstrations and milestone checkers are available for all ten tasks, and each task has a complete instruction sequence including required closing actions. Subgoal counts refer to distinct instruction phases in the execution plan, rather than the number of evaluation predicates. For the main language-supervision comparison, each condition evaluates 24 episodes per task and seed, giving 240 episodes per seed and 480 across seeds 42 and 142, with a 6,000-step limit. Performance is measured by strict task success, including required terminal conditions such as closing a cabinet; intermediate milestone completion alone is insufficient.

\begin{table}[htbp]
\centering
\small
\caption{RoboCasa 365 evaluation tasks by execution-plan depth. Subgoals count distinct instruction phases, not evaluation predicates.}
\label{tab:robocasa-task-depth}
\begin{tabular}{lr}
\toprule
Task & Subgoals \\
\midrule
BreadAndCheese & 2 \\
ThawInSink & 2 \\
StartElectricKettle & 2 \\
ClearCuttingBoard & 3 \\
GatherCuttingTools & 3 \\
RinseBowls & 3 \\
CerealAndBowl & 4 \\
MakeFruitBowl & 4 \\
RestockBowls & 4 \\
PlaceFoodInBowls & 4 \\
\bottomrule
\end{tabular}
\end{table}

\textbf{Instructor conditions.} Oracle evaluation uses authored subgoal instructions and simulator-state milestones for switching. Frontier-instructor evaluation separates a fixed initial plan from observation-based completion monitoring, as described in Appendix~\ref{app:frontier-instructor}. The planner identity and the policy checkpoint must be matched to the recorded run when interpreting a table cell.

\textbf{BEHAVIOR-1K evaluation.} For the main language-supervision comparison, we evaluate each condition for eight episodes on each of seven tasks per seed, totaling 56 episodes per seed and 112 across seeds 42 and 142. The seven tasks are Chop an onion, Hiding Easter eggs, Preparing lunch box, Setting mousetraps, Chopping wood, Sorting vegetables, and Putting away Halloween decorations. All rollouts use task instance 301, with each rollout executed in a separate process. The execution limit is 12,000 simulator steps per episode. We compute the mean terminal partial-credit score $q$ within each task, then report the unweighted mean of these seven task means. These are repeated rollouts of one instance per task, not eight distinct evaluation instances; they therefore do not establish generalization across a range of held-out task instances. The score measures benchmark goal-predicate satisfaction, not progress through planner-generated steps.

\textbf{Training configuration.} The policies in the simulation benchmark comparisons are trained for 120,000 steps. Crossword policy training uses a separate configuration described in Appendix~\ref{app:crossword-results}. The main language-supervision comparison uses two independently trained policies per condition, with training seeds 42 and 142; each is evaluated using the corresponding evaluation seed. The RoboCasa OpenPI configurations use a $\pi_{0.5}$ policy with an action horizon of ten, a maximum text length of 384 tokens, and batch size 256. Optimization uses AdamW with $(\beta_1,\beta_2)=(0.9,0.95)$, $\epsilon=10^{-8}$, weight decay $10^{-10}$, and gradient-norm clipping at 1.0. The learning rate warms up over 1,000 steps to $5\times10^{-5}$ and remains at that value; exponential moving average decay is 0.999. Because training and evaluation seeds change together, differences between the two runs reflect both sources of variation rather than isolating training-seed variability.

The BEHAVIOR-1K policy training configuration uses $\pi_{0.5}$ with an action horizon of 32, a maximum text length of 256 tokens, and batch size 64. Its learning-rate schedule warms up over 1,000 steps to $2.5\times10^{-5}$ and decays to $2.5\times10^{-6}$.

\textbf{Language supervision recipes.} Table~\ref{tab:supervision-recipes} specifies the prompt construction for each condition. Here, \texttt{goal} is the episode-level task instruction and \texttt{subgoal} is the current subtask instruction. In the hierarchical conditions, \texttt{hierarchical instruction} denotes the instruction corresponding to the sampled segment's hierarchy level, including primitive instructions such as ``lift the arm''. The prompt pairs this instruction with the original episode-level goal using the fixed phrase ``Current step:''. Multi-aspect supervision samples from the original segment instruction and descriptions in the four formats introduced in Section 4: arm/gripper kinematics, spatial destinations, terse commands, and natural phrasing.

\begin{table}[htbp]
\centering
\footnotesize
\setlength{\tabcolsep}{3pt}
\renewcommand{\arraystretch}{1.15}
\caption{Prompt construction for the six language supervision conditions. Both hierarchical conditions pair the episode-level goal with the instruction corresponding to the sampled segment's hierarchy level, using ``Current step:'' as a fixed phrase.}
\label{tab:supervision-recipes}
\begin{tabularx}{\linewidth}{@{}>{\hsize=.8\hsize\linewidth=\hsize\raggedright\arraybackslash}X>{\hsize=1.1\hsize\linewidth=\hsize\raggedright\arraybackslash}X>{\hsize=1.1\hsize\linewidth=\hsize\raggedright\arraybackslash}X@{}}
\toprule
Condition & Annotation levels in prompt & Training format \\
\midrule
Task-level & Episode-level goal only & \texttt{\{goal\}} \\
Subtask-level & Current subgoal & \texttt{\{subgoal\}} \\
Task + Subtask & Episode-level goal + current subgoal (no primitive supervision) & \texttt{\{goal\}}. \texttt{\{subgoal\}} \\
Hierarchical only & Episode-level goal + corresponding hierarchy instruction (e.g., primitive: ``lift the arm'') & \texttt{\{goal\}.}\newline\texttt{Current step:}\newline\texttt{\{hierarchical instruction\}} \\
Multi-aspect only & Current subgoal & \texttt{\{sampled subgoal\}} \\
Hierarchical + Multi-aspect & Episode-level goal + corresponding hierarchy instruction & \texttt{\{goal\}.}\newline\texttt{Current step:}\newline\texttt{\{sampled hierarchical instruction\}} \\
\bottomrule
\end{tabularx}
\end{table}

The hierarchical-only and joint conditions use the same goal context and fixed phrase; the joint condition additionally varies the segment instruction through multi-aspect rephrasings.

\subsection{Language Supervision Results by Seed}
\label{app:language-results-seeds}

Table~\ref{tab:language-results-seeds} reports the two seed results underlying the averages in Table~\ref{tab:language-results}. Each displayed mean is calculated before rounding. For RoboCasa, the mean is $(s_{42}+s_{142})/480$, where $s_{42}$ and $s_{142}$ are success counts over 240 trials per seed.

\begin{table}[htbp]
\centering
\footnotesize
\setlength{\tabcolsep}{3pt}
\renewcommand{\arraystretch}{1.35}
\caption{Language supervision results: mean, followed by (seed 42 / seed 142). Metrics are success rate for RoboCasa 365 and mean subgoal progress for BEHAVIOR-1K. The task-level condition uses a fixed task instruction and is shown once per benchmark.}
\label{tab:language-results-seeds}
\begin{tabularx}{\linewidth}{@{}>{\raggedright\arraybackslash}Xcccc@{}}
\toprule
 & \multicolumn{2}{c}{RoboCasa 365} & \multicolumn{2}{c}{BEHAVIOR-1K} \\
\cmidrule(lr){2-3}\cmidrule(lr){4-5}
Language Supervision & Oracle & Fable 5 & Oracle & Fable 5 \\
\midrule
Task-level (fixed) & \multicolumn{2}{c}{\shortstack{0.169\\(0.175 / 0.163)}} & \multicolumn{2}{c}{\shortstack{0.076\\(0.050 / 0.102)}} \\
Subtask-level & \shortstack{0.406\\(0.408 / 0.404)} & \shortstack{0.381\\(0.392 / 0.371)} & \shortstack{0.143\\(0.130 / 0.156)} & \shortstack{0.084\\(0.070 / 0.098)} \\
Task + Subtask & \shortstack{0.417\\(0.425 / 0.408)} & \shortstack{0.402\\(0.408 / 0.396)} & \shortstack{0.154\\(0.140 / 0.168)} & \shortstack{0.091\\(0.080 / 0.102)} \\
Hierarchical only & \shortstack{0.456\\(0.463 / 0.450)} & \shortstack{0.440\\(0.446 / 0.433)} & \shortstack{0.143\\(0.130 / 0.156)} & \shortstack{0.084\\(0.070 / 0.098)} \\
Multi-aspect only & \shortstack{0.425\\(0.433 / 0.417)} & \shortstack{0.406\\(0.417 / 0.396)} & \shortstack{0.160\\(0.150 / 0.170)} & \shortstack{0.103\\(0.090 / 0.116)} \\
Hierarchical + Multi-aspect & \shortstack{0.517\\(0.525 / 0.508)} & \shortstack{0.485\\(0.492 / 0.479)} & \shortstack{0.174\\(0.160 / 0.188)} & \shortstack{0.143\\(0.130 / 0.156)} \\
\bottomrule
\end{tabularx}
\end{table}

\section{Frontier Planning and Observation-Based Instruction Switching}
\label{app:frontier-instructor}

This appendix describes the frontier-instructor protocol used in Section 4 and the high-level plan evaluation in Probe 2. The protocol separates initial plan generation, execution monitoring, and instruction switching.

\subsection{Plan Generation and Reuse Across Policy Variants}

At the start of each episode ($t=0$), a frontier vision-language model receives the composite task description and the reset-frame camera views. It generates the complete ordered subgoal sequence once. For a given evaluation episode and planner/phrasing condition, this sequence is reused across all policy variants. Instruction content is therefore held fixed across policy variants while the language-supervision condition varies, following the same discipline as the oracle condition.

The frontier model supplies the plan before execution. Subsequent observations determine when the harness advances through that plan; they do not trigger regeneration of the subgoal sequence.

\subsection{Local Progress Monitoring}

A separate local open-weights vision-language model, Qwen3.6-27B, checks progress every 40 control steps using the active subgoal instruction and three-camera RGB observations from the subgoal's start and the current state. Its decisions use visual observations rather than simulator-state milestones. Monitor queries are synchronous.

The oracle condition instead uses an authored subgoal sequence and simulator-state milestones to determine completion. The frontier-instructor condition uses neither simulator state for initial planning nor simulator state for progress monitoring.

\subsection{Instruction Switching for the Local Policy}

The harness passes one instruction at a time from the pre-written subgoal list to the local policy. The policy executes the current instruction between monitor queries. Two consecutive positive judgments trigger advancement to the next instruction after the configured switching buffer. Uncompleted subgoals retain the current instruction without automatic skipping or replanning. Episodes terminate on benchmark success, environment termination, or exhaustion of the 6,000-step budget for RoboCasa 365 or the 12,000-step budget for BEHAVIOR-1K.

Thus, the instruction supplied to the local policy changes in response to observed completion, while the content and order of the subgoal list remain fixed. Plans are shared across policy variants, but switching times depend on each policy's observed execution progress.

\subsection{Connection to Probe 2: Robustness to Instruction Diversity}

The high-level plan evaluation in Section 4, Probe 2, evaluates the initial plan-generation stage described above. Opus 5, Gemini-3.1-Pro, and Fable 5 each generate a full subgoal sequence from the composite goal text and reset-frame camera views. Their plans are compared with oracle reference plans across 30 scenes using decomposition step-count match and semantic similarity, with Fable 5 serving as the semantic-similarity judge. Fable 5 compares individual generated steps with oracle-reference steps and directly assigns a semantic-similarity score in $[0,1]$ to each comparison. Higher scores indicate closer semantic agreement.

\begin{table}[htbp]
\caption{Plan quality against oracle references: step-count match and mean semantic similarity from direct 0--1 judgments of individual steps by Fable 5.}
\label{tab:planner-quality}
\centering
\small
\setlength{\tabcolsep}{3pt}
\renewcommand{\arraystretch}{1.05}
\begin{tabular}{@{}lccc@{}}
\toprule
Metric & Opus 5 & \shortstack{Gemini-\\3.1-Pro} & Fable 5 \\
\midrule
Step-count match & 29/30 & 25/30 & 27/30 \\
Mean semantic similarity & .886 & .855 & .895 \\
\bottomrule
\end{tabular}
\end{table}

The subsequent execution evaluation varies the source planner and whether its instructions are rephrased. Table~\ref{tab:planner-robustness} reports RoboCasa 365 results from one training/evaluation seed (42). Its original-instruction Fable 5 results match the seed-42 entries in Table~\ref{tab:language-results-seeds}, whereas Table~\ref{tab:language-results} averages seeds 42 and 142. Fable 5 generates the rephrased instructions for all three source planners. It is prompted to preserve the original task goal while changing the wording. For example, ``Open up the cabinet'' may be rewritten as ``Grasp the cabinet door handles and swing the doors open.'' Such rewrites can make action details explicit, so this condition tests variation in wording and descriptive detail rather than lexical substitution alone. Within each planner and phrasing condition, the generated sequence is reused across policy variants. All policy variants use the same local progress-monitoring and instruction-switching protocol. The plan-quality evaluation assesses the initial frontier-generated text; the execution evaluation measures how the policies perform with that text under observation-based instruction switching.

\section{Primitive Instruction Compliance}
\label{app:primitive-compliance}
Table~\ref{tab:primitive-compliance-full} reports the full primitive-compliance probe, including the instruction wording and movement threshold for each action.

We compare one checkpoint per condition from a single training seed of the subtask-level and hierarchical-only policies used in the main evaluation, without additional training for this probe.

Each command is evaluated three times in each of the 12 scenes, with a 24-second execution window per repetition. We average the three repetitions within each scene and then average equally across the 12 scenes.
Each repetition is scored as compliant when final displacement along the commanded axis exceeds its threshold. Thus, each reported rate aggregates 36 binary trials per command and policy.

\begin{table}[htbp]
\caption{Single-seed policy compliance rates for primitive kinematic instructions.}
\label{tab:primitive-compliance-full}
\centering
\small
\setlength{\tabcolsep}{3pt}
\renewcommand{\arraystretch}{1.15}
\begin{tabularx}{\linewidth}{@{}>{\hsize=1.65\hsize\linewidth=\hsize\raggedright\arraybackslash}X>{\hsize=0.7\hsize\linewidth=\hsize\raggedright\arraybackslash}X>{\hsize=0.95\hsize\linewidth=\hsize\raggedright\arraybackslash}X>{\hsize=0.7\hsize\linewidth=\hsize\raggedright\arraybackslash}X@{}}
\toprule
Primitive Action & Subtask-level & Hierarchical only & threshold \\
\midrule
Object up ("lift the object up") & 0.250 & 0.833 & 16~cm \\
Object down ("lower the object down") & 0.139 & 0.750 & 16~cm \\
Arm up ("move the arm up") & 0.111 & 0.750 & 34~cm \\
Arm down ("move the arm down") & 0.111 & 0.778 & 34~cm \\
Arm forward ("move the arm forward") & 0.056 & 0.556 & 24~cm \\
Arm back ("move the arm back") & 0.028 & 0.444 & 24~cm \\
\bottomrule
\end{tabularx}
\end{table}

\section{Crossword Evaluation Results}
\label{app:crossword-results}
Table~\ref{tab:crossword-per-game} reports individual game scores and completed-word counts underlying the means in Section~\ref{sec:crossword}. Each game uses a different alphabet set and a one-hour execution budget, with scoring as defined in the main text. The harness baseline uses subtask-level supervision, while the dense-supervision harness uses the hierarchical + multi-aspect recipe. Within each instruction condition, both use the same demonstrations, policy architecture, training budget, and GPT-6 Astra frontier controller; the comparison varies language supervision only. Word validity is judged by the authors based on familiar English usage. This is human adjudication rather than an automated dictionary-based metric. Repositioning an already credited cube or repeating an already credited word earns no additional points.

\textbf{Policy training configuration.} Both crossword policies are initialized from pretrained $\pi_{0.5}$~\citep{black2025pi05} and trained with an action horizon of 10, batch size 32, and a budget of 8,000 steps. The peak learning rate is $2.5\times10^{-5}$. These settings are shared by the subtask-level and hierarchical + multi-aspect supervision conditions.

\begin{table}[htbp]
\centering
\small
\setlength{\tabcolsep}{4pt}
\caption{Crossword results after one hour per game. Each game cell reports total score / completed words. Scores award $+2$ per correctly placed cube and $+2$ per completed valid word; the final column reports mean total score. The final two rows prohibit cell IDs in frontier-generated instructions.}
\label{tab:crossword-per-game}
\begin{tabularx}{\linewidth}{Xrrrrrr}
\toprule
Configuration & Game 1 & Game 2 & Game 3 & Game 4 & Game 5 & \shortstack{Mean\\score} \\
\midrule
Fable 5 (direct control) & 6 / 0 & 4 / 0 & 4 / 0 & 6 / 0 & 4 / 0 & 4.8 \\
GPT-6 Astra (direct control) & 8 / 0 & 4 / 0 & 10 / 1 & 10 / 1 & 8 / 0 & 8.0 \\
GPT-6 Astra harness & 14 / 2 & 16 / 2 & 14 / 1 & 14 / 1 & 14 / 1 & 14.4 \\
GPT-6 Astra harness + dense supervision & 16 / 2 & 16 / 2 & 14 / 1 & 16 / 1 & 16 / 2 & 15.6 \\
\midrule
\multicolumn{7}{l}{\textit{Cell IDs prohibited in instructions}} \\
GPT-6 Astra harness & 4 / 0 & 4 / 0 & 2 / 0 & 2 / 0 & 2 / 0 & 2.8 \\
GPT-6 Astra harness + dense supervision & 6 / 1 & 4 / 0 & 8 / 1 & 6 / 1 & 4 / 0 & 5.6 \\
\bottomrule
\end{tabularx}
\end{table}

\textbf{Instructions without cell IDs.} In an additional comparison, we explicitly instruct the frontier model not to use board cell IDs such as ``1A'', ``3B'', or ``6F'' in its directives to the local policy. It must instead describe destinations spatially, for example, ``put the cube on the left of the B cube'' rather than ``put the cube on 1A''. Table~\ref{tab:crossword-per-game} reports five games under this restriction. Mean scores fall from 14.4 to 2.8 without dense supervision and from 15.6 to 5.6 with dense supervision. Under the restriction, dense supervision yields higher scores in four games and ties in one. These results support improved performance with spatially described targets, while both policies remain substantially less effective when cell IDs are unavailable in instructions. Qualitatively, placement was more difficult because spatial descriptions could leave the intended target ambiguous; placement and recovery success were not separately quantified.

\textbf{Pick-and-place evaluation.} Table~\ref{tab:crossword-pick-place} reports accuracy over 20 trials per configuration, excluding retries. This measures first-attempt execution separately from the one-hour game scores above. The local policy can be evaluated on individual pick-and-place trials even though it cannot independently perform the full crossword task.

\begin{table}[htbp]
\centering
\small
\caption{Crossword pick-and-place accuracy over 20 trials per configuration, excluding retries.}
\label{tab:crossword-pick-place}
\begin{tabular}{lrr}
\toprule
Configuration & Successful trials & Accuracy \\
\midrule
Fable 5 (direct control) & 5/20 & 25\% \\
GPT-6 Astra (direct control) & 13/20 & 65\% \\
Local policy only & 10/20 & 50\% \\
GPT-6 Astra harness & 10/20 & 50\% \\
\bottomrule
\end{tabular}
\end{table}

\subsection{Observation and Execution Timing}

The frontier model autonomously decides when to read the available camera views and how long to execute each action segment. For example, it can select 2--3\,s segments for transport and 0.12\,s segments for fine adjustments. These durations describe physical execution rather than frontier-model inference latency.

\begin{figure}[htbp]
\centering
\includegraphics[width=\linewidth]{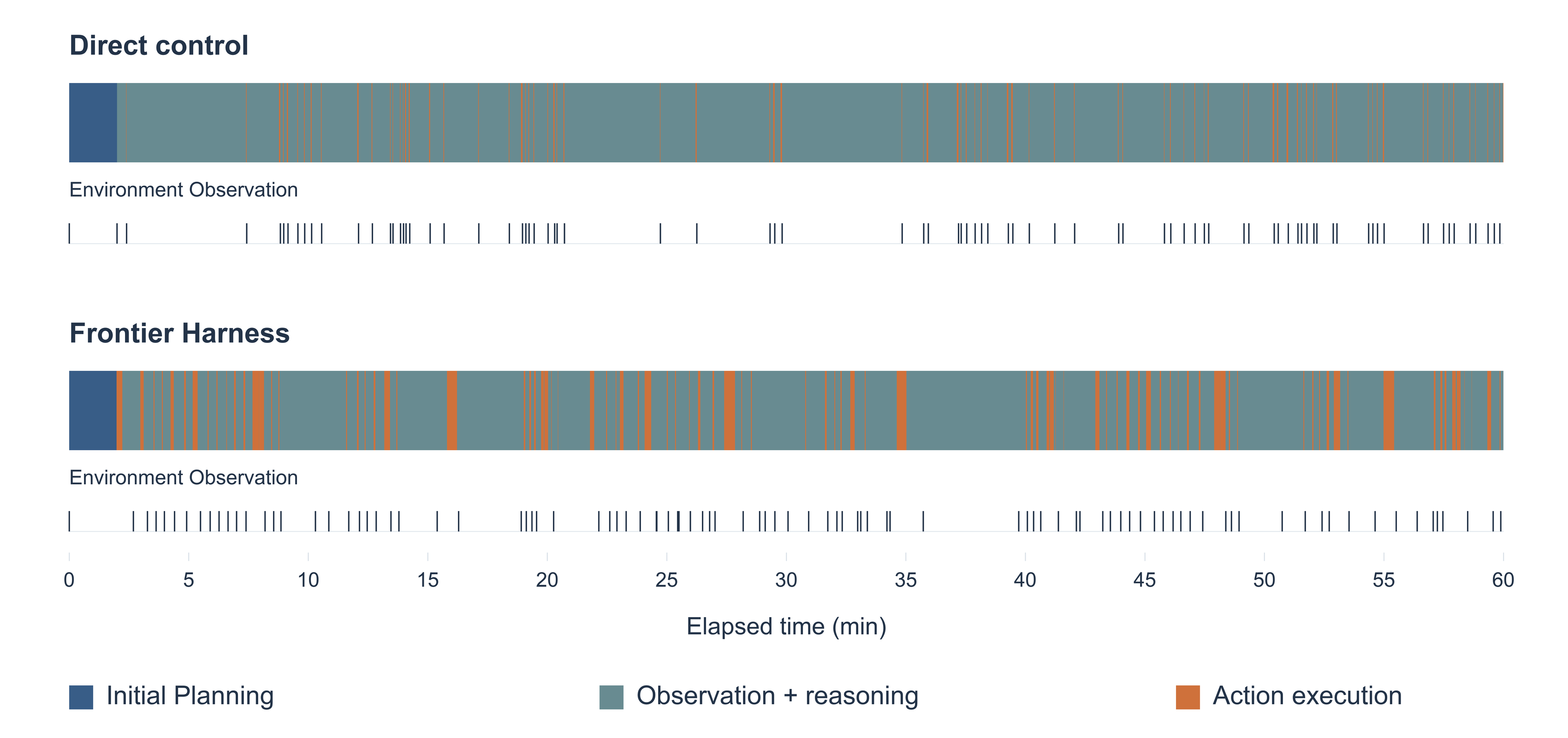}
\caption{Full 60-minute crossword activity timelines for GPT-6 Astra, one run per configuration. Ticks mark camera observations requested by the model. Cumulative action-execution time is 2 min 40 s under direct control and 9 min 8.55 s with the harness; these are individual runs, not averages across games.}
\label{fig:crossword-activity-full}
\end{figure}

Figure~\ref{fig:crossword-activity-full} compares one-hour activity timelines for GPT-6 Astra under direct control and frontier harness execution. In the illustrated direct-control run, GPT-6 Astra repeatedly reasons about successive raw actions, with 2 min 40 s spent on physical execution. In the illustrated harness run, action generation is delegated to the local policy, and physical execution occupies 9 min 8.55 s. This example is consistent with the harness allocating more time to robot motion while retaining observation-based guidance and correction. These durations correspond to 4.4\% and 15.2\% of the one-hour budget, respectively, or 3.43 times as much physical execution with the harness. This ratio measures time spent executing actions, not movement speed; the comparison uses one illustrated run per configuration and does not estimate an average timing advantage across games.

\end{document}